%% file: main.tex
\documentclass{article}

\usepackage[preprint]{neurips_2025}

\usepackage[utf8]{inputenc} 
\usepackage[T1]{fontenc}    
\usepackage{hyperref}       
\usepackage{url}            
\usepackage{booktabs}       
\usepackage{amsfonts}       
\usepackage{nicefrac}       
\usepackage{microtype}      
\usepackage[table,xcdraw,dvipsnames,svgnames]{xcolor} 
\usepackage{pifont}
\usepackage{amssymb}
\usepackage{bbding}
\usepackage{bm}

\usepackage{makecell}
\usepackage{wrapfig}
\usepackage{algorithm}
\usepackage{algpseudocode}
\usepackage{amsmath}
\usepackage{amssymb}
\usepackage{mathtools}
\usepackage{amsthm}
\usepackage{bm}
\usepackage[capitalize,noabbrev]{cleveref}
\usepackage{enumitem}
\usepackage{pifont}         
\usepackage{multirow,multicol,makecell}
\usepackage{colortbl}
\usepackage{tcolorbox}
\definecolor{light-gray}{gray}{0.94}

\newcommand{\xhdr}[1]{\noindent{{\bf #1.}}}

\title{
Enabling Flexible Multi-LLM Integration for Scalable Knowledge Aggregation
}

\author{
    Zhenglun Kong$^{1,3}$\thanks{Equal contribution.},
    Zheng Zhan$^1$\footnotemark[1],
  Shiyue Hou$^1$, Yifan Gong$^1$, Xin Meng$^2$, \\
  \textbf{Pengwei Sui$^3$, Peiyan Dong$^1$, Xuan Shen$^1$, Zifeng Wang$^4$,} \\
  \textbf{Pu Zhao$^1$, Hao Tang$^2$, Stratis Ioannidis$^1$, Yanzhi Wang$^1$}\\
  $^1$Northeastern University, 
  $^2$Peking University,
  $^3$Harvard University,
  $^4$Google \\
}

\begin{document}

\maketitle

\begin{abstract}
Large language models (LLMs) have shown remarkable promise but remain challenging to continually improve through traditional finetuning, particularly when integrating capabilities from other specialized LLMs. 
Popular methods like ensemble and weight merging require substantial memory and struggle to adapt to changing data environments.  
Recent efforts have transferred knowledge from multiple LLMs into a single target model; however, they suffer from interference and degraded performance among tasks, largely due to limited flexibility in candidate selection and training pipelines.  
To address these issues, we propose a framework that adaptively selects and aggregates knowledge from diverse LLMs to build a single, stronger model, avoiding the high memory overhead of ensemble and inflexible weight merging. 
Specifically, we design an adaptive selection network that identifies the most relevant source LLMs based on their scores, thereby reducing knowledge interference.   
We further propose a dynamic weighted fusion strategy that accounts for the inherent strengths of candidate LLMs, along with a feedback-driven loss function that prevents the selector from converging on a single subset of sources.
Experimental results demonstrate that our method can enable a more stable and scalable knowledge aggregation process while reducing knowledge interference by up to 50\% compared to existing approaches. \footnote{Code is avaliable at \href{https://github.com/ZLKong/LLM_Integration}{Github}.}

\end{abstract}

\input{1_introduction}

\input{2_related_work}

\input{3_method}
\input{4_results}
\input{5_ablation}

\input{6_conclusion}

\bibliographystyle{plain}
\bibliography{reference}


\newpage
\appendix
\input{7_appendix}



\end{document}

%% file: 1_introduction.tex
\section{Introduction}

The emergence of large language models (LLMs) has revolutionized various domains, catalyzing the development of many specialized models \cite{zhang2023alpacareinstructiontuned,christophe2024med42,koala_blogpost_2023,leng2023chinese-vicuna,li2023starcoder,li2025comprehensive,wang2025systematic,li2024distinct}. 
Training a new LLM from scratch is prohibitively expensive in terms of computational resources and data requirements. As a result, the prevalent strategy is to finetune existing models using localized or task-specific data.

While traditional finetuning can incrementally enhance a model’s performance within a specific domain, it often fails to leverage the rich, domain-specific expertise embedded in other LLMs, especially when relevant datasets are inaccessible or require extensive pre-processing~\cite{li2024towards}.
Furthermore, for organizations that have invested heavily in developing and tailoring their own LLMs, replacing the current model with a new, pre-trained one introduces challenges such as re-adaptation, retraining, and potentially significant costs to maintain alignment with their unique requirements.
Therefore, we focus on building a stronger LLM by incorporating knowledge from multiple specialized models. Instead of finetuning a single model in isolation, we aggregate knowledge from various LLMs to enhance performance in a stable, efficient, and scalable way.  This approach preserves the strengths of the existing model while infusing it with complementary knowledge, ensuring that the final model is both highly capable and aligned with specific needs.

\begin{wrapfigure}{t}{0.6\textwidth}
\vspace{-0.1cm}
  \centering
  \includegraphics[width=0.6\textwidth]{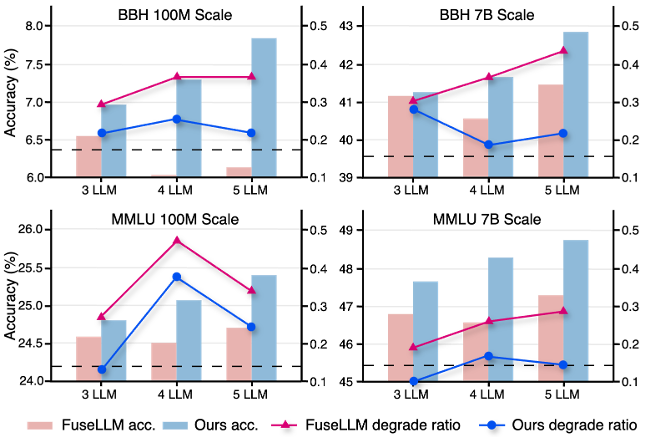}
    \vspace{-0.5cm}
  \caption{\textbf{Scaling number of fusion candidates.} We show accuracy (histogram, higher the better) and the percentage of tasks degrading the baseline (Line chart, lower the better) when integrating three, four, and five LLMs on the BBH and MMLU benchmarks. Dotted lines represent the baseline.}
  \label{fig:bbh_trend_candidiate_selection}
  \vspace{-0.2cm}
\end{wrapfigure}
Existing solutions, such as ensemble methods \cite{jiang2023llm,lu2023routing,zhong2025enhancing}, enhance prediction performance by aggregating outputs from multiple models but require additional memory and increased inference time due to operating several models simultaneously. Another method involves merging several neural networks into a single network within the parameter space \cite{jin2022dataless}. This generally requires ensuring uniform architecture and depends on manually configured weight merging or adding additional layers.
Additionally, Mixture of Expert (MoE) structures
such as Mistral-7bx8 \cite{jiang2024mixtral}, address some inference and weight-sharing issues, but still face long inference times, homogeneous architectures, and larger model sizes.
FuseLLM~\cite{wan2024knowledge} and FuseChat~\cite{wan2024fusechat} attempted to integrate the knowledge of multiple source LLMs using generated probability distribution matrices. However, these approaches suffer from interference and performance degradation in various tasks compared to the original target model due to suboptimal model selection and uncontrolled fusion processes.

To overcome the limitations of existing LLM integration approaches, we propose a dynamic framework that adaptively selects LLMs for integration. Given a diverse set of source LLMs with heterogeneous structures, we introduce an adaptive selection network, a learnable mechanism that explicitly evaluates and selects the best-performing source LLMs based on their important scores, thereby alleviating interference issues typically associated with model fusion. 
The scores are computed based on the performance of each model across a predefined set of tasks. Our framework provides flexibility in the number of LLMs selected during this process.

To improve the knowledge aggregation process, we propose a dynamic weighted fusion strategy that considers the intrinsic characteristics of candidate LLMs during fusion. The assigned weights are derived from the score evaluations, allowing the fusion process to prioritize models that are more likely to enhance the overall performance of the composite LLM.
The selector often converges to a state in which it consistently assigns large weights to a small subset of candidates. To mitigate this, we introduce a feedback-driven loss function that optimizes the training of our adaptive selection network and guides the selection of candidates.

Our method ensures stable and scalable integration of LLMs while maintaining both efficiency and effectiveness despite model diversity, as shown in Fig.~\ref{fig:bbh_trend_candidiate_selection} (Detailed analysis in Sec.~\ref{sec:abl}: Model Scaling Results). It achieves this without increasing the parameter size or computation
of the target model, making it more efficient compared to traditional methods.
Our contributions are as follows:

\begin{itemize}[leftmargin=*, noitemsep, topsep=0pt]
\item We find that merely increasing the number of fusion candidates and expanding the source model pool does not necessarily enhance the fusion process, a selective strategy is more effective in minimizing knowledge interference.
\item We propose a novel dynamic integration framework that adaptively selects LLMs for integration, leveraging an adaptive selection network, a dynamic weighted fusion strategy, and a feedback-driven loss function to alleviate interference issues and enhance performance.
\item Our model shows improvement in accuracy across multiple benchmarks as more models are integrated, while reducing knowledge interference by up to 50\% compared to previous methods.
\end{itemize}

%% file: 2_related_work.tex
\section{Related Work}

\xhdr{Model Integration} Research on model integration has evolved into distinct categories, each addressing different aspects of combining models \cite{wang2024learn,wang2024rehearsal}:
 1) \textit{Ensemble:} LLM-Blender \cite{jiang2023llm} 
uses ensemble techniques to enhance performance by combining outputs from multiple models. 
\begin{figure*}[t]
\vspace{-0.3cm}
  \centering
  \includegraphics[width=1.0\linewidth]{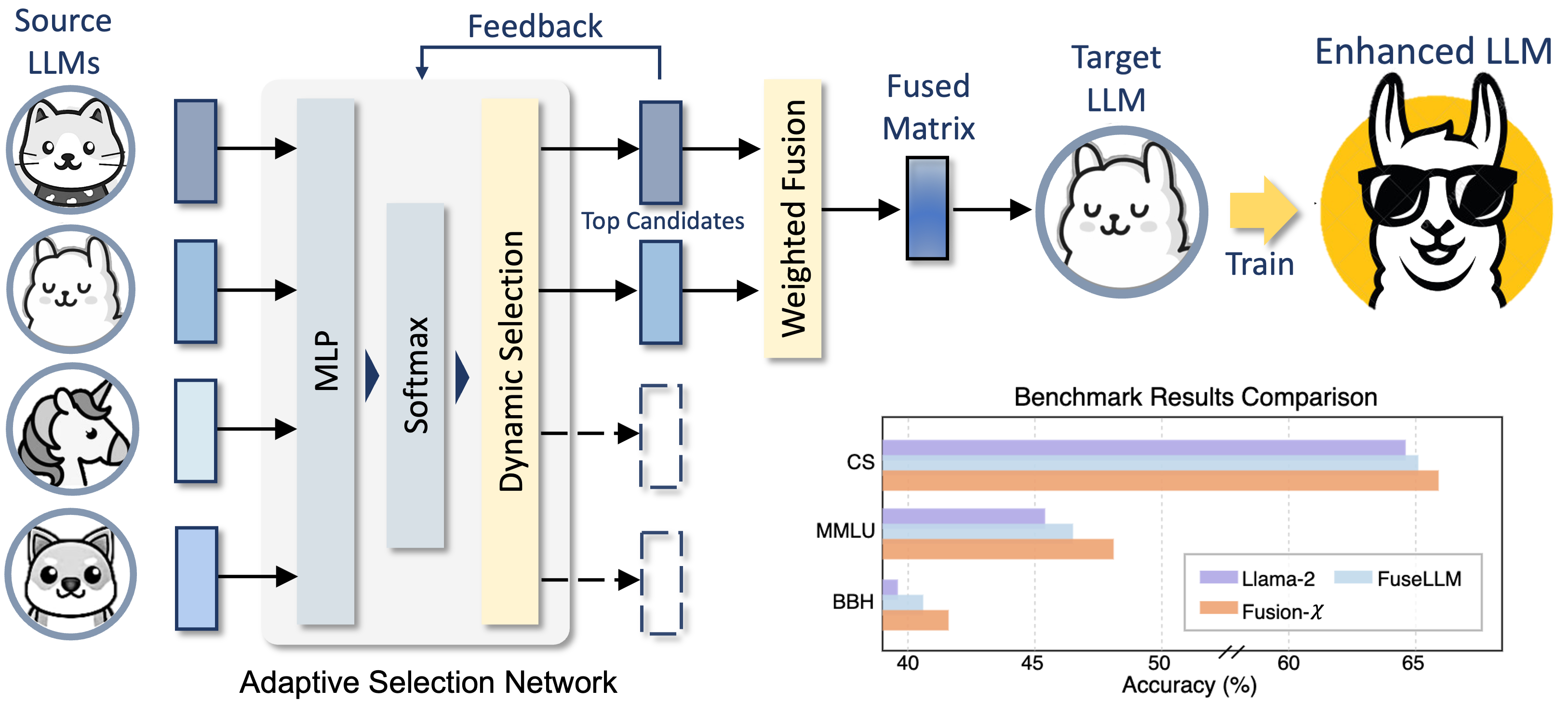}
  \vspace{-0.2cm}
  \caption{
  \textbf{Overall framework:} Multiple LLMs are evaluated and selected based on performance by an adaptive selection network. Top candidates then proceed through a dynamic weighted fusion process guided by a feedback loss to enhance the ability of the target LLM. The lower right shows results on CommonSense, MMLU, and Big-Bench Hard benchmark.}
  \label{fig:main_framework}
  \vspace{-0.15cm}
\end{figure*}
This process includes inferring all candidate 
models and then ranking them, which can be resource-intensive and slow.
2) \textit{Weight Merging:} Zipit \cite{stoica2023zipit} merges partial layers of two models without additional training, creating a multi-head model for various tasks.
\cite{rame2022diverse,arpit2022ensemble,wortsman2022model} employ weighted averaging methods. 
TIES-Merging \cite{yadav2024ties} eliminates parameter interference among multiple models by removing delta parameters with low magnitudes and merging parameters with consistent signs. 
\cite{zhang2023composing} compose models through linear arithmetic operations in the weight space. These techniques are typically limited to models with identical architectures.
3) \textit{Knowledge Fusion:} FuseLLM \cite{wan2024knowledge} and FuseChat \cite{wan2024fusechat} focuses on fusing the probability distributions from various LLM candidates, integrating them into a single base LLM, blending knowledge across models.  
Knowledge distillation \cite{hinton2015distilling} is also used to integrate information into a model. However, student models are typically smaller and have lower performance than their teacher models. In our scenario, there is no limitation on the size or performance of the source models.

%% file: 3_method.tex
\section{Preliminaries and Motivation}
\label{sec:pre}

\xhdr{Preliminaries} As a general integration approach parallel to ensembling and weight merging, knowledge fusion \cite{wan2024knowledge} combines the probabilistic distribution matrices from a set of $M$ LLMs, denoted as $\{P^{\theta_i}_t\}^M_{i=1}$, where $\theta_i$ represents the parameters of the $i$-th LLM.  These reflect each model's inherent knowledge for text understanding.
Let $t$ be a text sequence of length $N$, and $t_{<n}$ denote the sequence preceding the $n$-th token. The probabilistic distribution matrix $\textbf{P}_t^{\theta_i}$ for the $i$-th LLM is:
\begin{equation} 
\textbf{P}_t^{\theta_i} = \left[ \textbf{p}_{\theta_i}(t_1|t_{<1}), \textbf{p}_{\theta_i}(t_2|t_{<2}), \ldots, \textbf{p}_{\theta_i}(t_n|t_{<n}) \right],
\end{equation} 
where $\textbf{p}_{\theta_i}(t_n|t_{<n})$ is the predicted probability distribution for the $n$-th token given the preceding tokens $t_{<n}$, according to the $i$-th LLM parameterized by $\theta_i$.
Each element $\textbf{p}_{\theta_i}(t_k|t_{<k})$ is a vector of probabilities corresponding to each token in the vocabulary, summing to 1.

Their fusion process is achieved by minimizing the divergence between the probabilistic distributions of target LLM (pre-defined among the source LLMs) and source LLMs:
\begin{equation}
\textbf{P}_{f} = \mathcal{F}(\textbf{P}_{t}^{\theta_1}, \textbf{P}_{t}^{\theta_2}, \ldots, \textbf{P}_{t}^{\theta_M}),
    \label{eq:prob_matrix}
\end{equation} 
where $\mathcal{F}$ is the function that combines multiple matrices. 

The overall objective for continual training consists of a weighted combination of the causal language modeling and the fusion objective:
\begin{equation}
\mathcal{L} = \lambda \mathcal{L}_{\text{lm}} + (1 - \lambda) \mathcal{L}_{\text{fuse}},
\label{eq:total_loss}
\end{equation}
where $\mathcal{L}_{\text{lm}}$ is the causal language modeling objective, and $\mathcal{L}_{\text{fuse}}$ is the cross-entropy loss between the target LLM’s predictions (output) and the fused representation matrix $\textbf{P}_f$. 

\xhdr{Motivation} 
We conduct an evaluation of FuseLLM \cite{wan2024knowledge} on 27 tasks of the Big-Bench Hard \cite{suzgun2022challenging} benchmark and 57 tasks from the Multi-task Language Understanding (MMLU), as shown in Fig. \ref{fig:bbh_trend_candidiate_selection}. 
There are two key observations: 
\begin{itemize}[leftmargin=*, noitemsep, topsep=0pt]
\item A high percentage of tasks exhibit performance degradation (red trend lines) compared to the original unmerged model (detailed numbers are shown in Tab. \ref{tab:beg_bench}).
\item Integrating more models leads to progressively greater performance degradation (red bars), emphasizing the impact of knowledge interference. 
\end{itemize}

This phenomenon can arise due to: 
1) \textit{Dilution of Valuable Knowledge}: Introducing less relevant or lower-quality information can dilute the original model's knowledge \cite{si2023knowledge}.
2) \textit{Overfitting to Irrelevant Patterns}: The fused model may overfit to noise or less useful patterns from new models, reducing its ability to perform the tasks it was originally trained for. 


\section{Methodology}
\label{sec:method}

Motivated by the above observations, we propose a fusion framework to advance existing knowledge fusion methods by introducing a dynamic framework that consists of an Adaptive Selection Network and a Dynamic Weighted Fusion
mechanism, as illustrated in Fig. \ref{fig:main_framework}.  Specifically, at each training step, the Adaptive Selector evaluates performance metrics to dynamically select a subset of candidate models based on their probabilistic distribution matrices rather than all candidates. 
More importantly, both the selection of candidates and the number of candidates selected are adaptive, preventing knowledge interference and enhancing overall model performance.
The selected candidates are then fused using a weighted sum based on normalized selection probabilities. 
This process is guided by a specially designed loss function that refines model selection through feedback. 
Our framework provides flexibility for future scalability and allows the integration process to accommodate varying computational constraints and application needs.

\subsection{Adaptive Selection Network}

We propose an Adaptive Selection Network (ASN) to evaluate the source models based on a continuous learning process. It integrates feedback from ongoing interaction, which will be introduced in Sec. \ref{sec:loss}. 
The network takes the normalized matrices \( \{\textbf{P}_t^{\theta_i}\}_{i=1}^{M} \) (simplified as $\textbf{P}_i$ in later equations) as input. These matrices are flattened and normalized using layer normalization to stabilize training. The network then computes the logits for each candidate model. 
 It consists of three linear layers with specified dimensions and uses the GELU activation function to introduce non-linearity, thereby enhancing its ability to capture complex patterns in the input data.  We concatenate all $\textbf{P}_i$ as $\textbf{P}_{cat}$ and send into the module to obtain logits:
\begin{equation} 
\textbf{z}_\phi = (f^3 \circ \text{GELU} \circ f^2 \circ \text{GELU} \circ f^1)\textbf{P}_{cat},
    \label{eq:gating}
\end{equation} %
where \( f^1 \), \( f^2 \), and \( f^3 \) represent linear layers. 
The adaptive selection mechanism utilizes the scores from the network, converting these into a probability via the softmax function $\textbf{p}_i = {e^{z_\phi}}/{\sum_{i=1}^N e^{z_\phi}},$

where \( \textbf{p}_i \) is the softmax probability associated with the \(i\)-th candidate. We also compared the effects of adding Gumbel softmax \cite{jang2016categorical} or noise \cite{shazeer2017outrageously} before the softmax (see Tab.~\ref{tab:ablation} Selection metric).
The better performance of softmax shows that the selection process benefits more from the smooth and differentiable mapping of logits, as well as improved convergence, rather than from adding randomness and increasing variance.

\paragraph{Dynamic Candidate Selection.}
To determine which candidate models to select for fusion, we apply a dynamic thresholding mechanism. Candidates with selection probabilities exceeding the threshold \( \tau \) are selected:
\begin{equation}
\small
\mathcal{X}_{\text{selected}} = \left\{ \textbf{P}_{t}^{\theta_j} \mid \textbf{p}_j > \tau, \, j = 1, \ldots, M \right\},
    \label{eq:threshold_selection}
\end{equation}
where the output set \( \mathcal{X}_{\text{selected}} = \{\textbf{P}_{t}^{\theta_j}\}_{j=1}^{K} \) represents a subset of the original set of $M$ LLMs. We simplify the notation $\textbf{P}_{t}^{\theta_j}$ as $\textbf{P}_{j}$ in the later equations. 

To ensure that at least one candidate is selected per sample $1 \leqslant K  \leqslant M$, we check if no candidates meet the threshold, then select the candidate with the highest probability:
\begin{equation}
\mathcal{X}_{\text{selected}} = \{ \arg\max_j \textbf{p}_j \}, \quad \text{if } |\mathcal{X}_{\text{selected}}| = 0,
\label{eq:ensure_selection}
\end{equation}
We set \( \tau = 0.15 \) in our implementation. This approach allows the model to adaptively choose the most relevant candidates based on input data and current learning context.

\input{subtables}
\subsection{Dynamic Weighted Fusion}

We proceed with the fusion process after selecting the candidate models. 
First, we normalize the weights of the selected probabilities $\textbf{p}_i$:
\begin{equation}
\hat{\textbf{p}} = \pi \left(\frac{\textbf{p} \odot m}{\sum_{i=1}^M \textbf{p}_im_i + \epsilon} \right),
\label{eq:normalized_probs}
\end{equation}
where \( m_i \) is a binary mask indicating the selected candidates (\( m_i = 1 \) if \( \textbf{p}_i \in \mathcal{X}_{\text{selected}} \), else \( m_i = 0 \)), and \( \epsilon \) is a small constant to prevent division by zero.  $\pi(\cdot)$ is a function to resize the vector by removing 0-valued elements, so that $\hat{p}$ has $K$ elements corresponding to $K$ selected LLMs despite of $M$ elements in $\textbf{p}$. 
To perform the weighted sum, we reshape the normalized probabilities and masks to match the dimensions of the candidate outputs, enabling element-wise multiplication. 
The outputs of the $K$ selected candidates $\textbf{p}_j$ are accumulated based on their respective weights to produce a unified model output $\textbf{P}_{f}$. This is calculated as follows:
\begin{equation}
    \textbf{P}_{f} = \texttt{sum} \left( \texttt{concat} \left( \left\{ \textbf{P}_j \cdot \hat{\textbf{p}_j}\right\}_{j=1}^{K},\ \texttt{dim=-1} \right), \ \texttt{dim=-1} \right),
    \label{eq:combined_assign_fuse}
\end{equation}
We assign the proportion of the candidates' probabilistic
distributions based on the weights in Eq.~\eqref{eq:normalized_probs}. 
$\texttt{concat}$ denotes the concatenation of all $K$-selected candidates.
$\texttt{sum}$ function is for a weighted sum of these aligned metrics $\textbf{P}_f$.
This dynamic fusion process can constantly let the more influential candidates have a greater effect on the final model. 
Next, the fused representation \( \textbf{P}_{f} \) goes through the cross-entropy loss $\mathcal{L}_{\text{fuse}}$.
We highlight that we explore various configurations for the fusion method,  threshold, and so on,  as shown in Tab.~\ref{tab:ablation}, and choose the configuration with the best performance.  
Our method fundamentally transforms the approach to integration by utilizing a data-driven, adaptive mechanism to dynamically evaluate contributions of candidate LLMs and select accordingly. 

\subsection{Loss and Training Pipeline}
\label{sec:loss}
In practice, we find that the selection network often converges to a state where it consistently assigns large weights to the same few candidates. To mitigate this issue, we implement a feedback approach to guide the selection of candidates. Consequently, we adopt a soft constraint approach. The importance of a model relative to a batch of training examples is defined as the batch-wise sum of the values $\hat{\textbf{p}_j}$ for each LLM.
We define a feedback loss \( L_{\text{feed}} \), which is added to the existing loss function for the model as described in Sec. \ref{sec:pre}. This loss is calculated as the square of the coefficient of variation $\mathcal{CV}^2$ of the importance values. The importance values are derived from the weights of different candidates in the model, summed over the index set \( K \). This formulation is given by:
\begin{equation}
\small
\mathcal{L}_{\text{feed}} = \mathcal{CV}^2\left(\{\hat{\textbf{p}_j}\}_{j=1}^{K} \right) = \frac{\sigma^2\left(\{\hat{\textbf{p}_j}\}_{j=1}^{K}\right)}{\mu^2\left(\{\hat{\textbf{p}_j}\}_{j=1}^{K}\right) + \epsilon},
\label{eq:square_loss}
\end{equation}%

Here, \( \sigma^2 \) is the variance, \( \mu \) is the mean, and \( \epsilon \) is a small constant added to ensure numerical stability. This refined definition emphasizes the goal of making the distribution of source LLMs' importance more uniform across the model. Minimizing the variance of the importance values \(\hat{\textbf{p}_j}\) reduces the spread or difference between these values, making the distribution of importance more uniform. Simultaneously maximizing the mean ensures that the feedback loss does not become excessively sensitive to small variances. Squaring the mean in the denominator helps to normalize the loss and maintain a consistent scale, emphasizing relative changes in the variance.
The full objective is:
\begin{equation}
\small
\mathcal{L} (\theta_{\mathcal{T}}, \phi_{ASN}) = \underbrace{-\mathbb{E}_{t \sim \mathcal{C}} \left[ \mathcal{D}(\mathcal{T}_t, O_t) \right]}_{\mathcal{L}_{\text{lm}}} 
+ \underbrace{\lambda_{\text{fuse}}\left(-\mathbb{E}_{t \sim \mathcal{C}} \left[ \mathcal{D}(\mathcal{T}_t, P_f) \right]\right)}_{\mathcal{L}_{\text{fuse}}}  + \underbrace{\lambda_{\text{feed}} \mathcal{CV}^2\left(\sum\nolimits_{j \in K} \hat{\textbf{p}_j}\right)}_{\mathcal{L}_{\text{feed}}},
\label{eq:full_training_loss}
\end{equation}%
where $\theta_{\mathcal{T}}, \phi_{ASN}$ are parameters of the target LLM and the selection network. $\mathcal{L}_{\text{lm}}$ reduces the discrepancy between target model output \( \mathcal{T}_t \) and the one-hot label matrix \( O_t \in \{0, 1\}^{N \times V} \), where V is the vocabulary size.
$\mathcal{L}_{\text{fuse}}$ enforces assignment between the output of the target LLM \( \mathcal{T}_t \) and the fused representation matrix \(P_f\).
We set \(\lambda_{\text{fuse}} = 0.1\) and \(\lambda_{\text{feed}} = 0.5\) in our experiments. Grid search results are shown in Appx. \ref{sup:train_detail}.   Our training algorithm is described in the Appx.~\ref{sup:design}. 

%% file: subtables.tex
\begin{table}[t]
\centering
\caption{\textbf{Different Design Choices} of our framework under Fusion-$\mathcal{X}$-T scale with four source models. We show ablation results on Commonsense (CS) and BBH, along with perplexity (PPL).}
\label{tab:ablation}
\begin{minipage}{0.48\textwidth}
\centering
\resizebox{1.0\linewidth}{!}{\begin{tabular}{c|l|c c c}
\toprule
\textbf{Category} & \textbf{Setting} & \textbf{PPL}$\downarrow$ & \textbf{CS}$\uparrow$ & \textbf{BBH}$\uparrow$ \\
\midrule
\multirow{3}{*}{\makecell[c]{\textbf{Selection}\\\textbf{count}}}
 & Top-2    & 11.67 & 40.55 & 6.75 \\
 & \cellcolor{blue!12}Adaptive  & \cellcolor{blue!12}\textbf{11.04} & \cellcolor{blue!12}\textbf{41.32} & \cellcolor{blue!12}\textbf{7.31} \\
 & All     & 11.91 & 40.52 & 6.64 \\
\midrule
\multirow{3}{*}{\makecell[c]{\textbf{Selection}\\\textbf{metric}}}
 & \cellcolor{blue!12}Softmax   & \cellcolor{blue!12}\textbf{11.04} & \cellcolor{blue!12}\textbf{41.32} & \cellcolor{blue!12}\textbf{7.31} \\
 & Gumbel   & 13.41 & 39.15 & 5.00 \\
 & Noise    & 13.11 & 38.97 & 4.90 \\
\midrule
\multirow{3}{*}{\makecell[c]{\textbf{Layer}\\\textbf{choice}}}
 & Conv.        & 12.21 & 39.73 & 6.11 \\
 & 1$\times$Linear & 11.42 & 40.78 & 6.86 \\
 & \cellcolor{blue!12}3$\times$Linear & \cellcolor{blue!12}\textbf{11.04} & \cellcolor{blue!12}\textbf{41.32} & \cellcolor{blue!12}\textbf{7.31} \\
\bottomrule
\end{tabular}}
\end{minipage}%
\hfill
\begin{minipage}{0.5\textwidth}
\centering
\resizebox{1.0\linewidth}{!}{\begin{tabular}{>{\centering\arraybackslash}p{1.4cm}|l|c c c}
\toprule
\textbf{Category} & \textbf{Setting} & \textbf{PPL}$\downarrow$ & \textbf{CS}$\uparrow$ & \textbf{BBH}$\uparrow$ \\
\midrule
\multirow{4}{*}{\makecell[c]{\textbf{Fusion}\\\textbf{method}}}
 & Avg            & 11.32 & 40.85 & 6.80 \\
 & Max            & 11.77 & 40.11 & 6.68 \\
 & w/o Weight.   & 11.96 & 39.82 & 6.38 \\
 & \cellcolor{blue!12}Weight.   & \cellcolor{blue!12}\textbf{11.04} & \cellcolor{blue!12}\textbf{41.32} & \cellcolor{blue!12}\textbf{7.31} \\
\midrule
\multirow{3}{*}{\makecell[c]{\textbf{Threshold}\\\textbf{setting}}}
 & 0.2   & 13.78 & 39.24 & 5.07 \\
 & \cellcolor{blue!12}0.15  & \cellcolor{blue!12}\textbf{11.04} & \cellcolor{blue!12}\textbf{41.32} & \cellcolor{blue!12}\textbf{7.31} \\
 & 0.12  & 11.67 & 40.65 & 6.75 \\
\midrule
\multirow{2}{*}{\makecell[c]{\textbf{Adding}\\\textbf{loss}}}
 & w/o Loss   & 11.48 & 40.91 & 6.92 \\
 & \cellcolor{blue!12}Feed. loss   & \cellcolor{blue!12}\textbf{11.04} & \cellcolor{blue!12}\textbf{41.32} & \cellcolor{blue!12}\textbf{7.31} \\
\bottomrule
\end{tabular}}
\end{minipage}
\end{table}

%% file: 4_results.tex
\vspace{-0.4cm}
\section{Experiments}
\subsection{Implementation Details}
\label{sec:datasets}
\textbf{Models and Datasets.} 
Following existing methods \cite{jiang2023llm,wan2024knowledge, goddard2024arcee, wang2023fusing}, we use llama-2-7B as the target model and evaluate on various benchmarks for fair comparison.
To demonstrate scaling performance of both parameter size and number of models, we evaluate on mutiple scales, including Llama-160M \cite{miao2023specinfer}, GPT-Neo-125M \cite{gpt-neo}, Pythia-160M \cite{biderman2023pythia}, Tiny-starcoder \cite{li2023starcoder}, LiteLlama-460M-1T, OpenLLaMA-V2-3B \cite{openlm2023openllama}, MiniMA-3B \cite{zhang2023law}, Amber\cite{liu2023llm360}, Starcoder2-3B \cite{li2023starcoder}, Llama-2-7B \cite{touvron2023llama}, OpenLLaMA-7B \cite{openlm2023openllama}, MPT-7B \cite{MosaicML2023Introducing}, Pythia-6.9B \cite{biderman2023pythia}, Starcoder2-7B~\cite{li2023starcoder}, Llama 3-8B \cite{grattafiori2024llama}, Yi-6B \cite{young2024yi}.
These models have different parameter sizes, architectures, tokenizers, and vocabularies. We follow~\cite{wan2024knowledge} to use MiniPile \cite{kaddour2023minipile} for continual training. 

\xhdr{Training details} Our model is optimized using the AdamW optimizer with beta1 = 0.9 and beta2 = 0.95, with gradient clipping set to 1.0 and weight decay to 0.1. A cosine learning rate schedule is employed, with a maximum learning rate of 3e-5 for models under 1B and 1e-5 for models larger than 1B and a warmup ratio of 0.008. We train with 8 A100 GPUs, each with 80GB of memory.

\textbf{Evaluation benchmarks.} We evaluate Fusion-$\mathcal{X}$ on three benchmarks that represent different core capabilities of LLMs: Common Sense (CS) \cite{talmor2018commonsenseqa}, Big-Bench Hard (BBH) \cite{suzgun2022challenging}, Multi-task Language Understanding (MMLU) \cite{hendrycks2021measuringmassivemultitasklanguage}, and MultiPL-E (ME) \cite{cassano2023multipl}, representing the ability of 
commonsense, reasoning, and code generation. 

\vspace{-0.1cm}
\subsection{Main Results} 
\label{sec:main_results}
\textbf{Common Sense (CS) Evaluation.} Tab. \ref{tab:commonsense} shows the zero-shot performance of Fusion-$\mathcal{X}$ and the baseline methods when fusing 4 LLMs. We present our model in three scales: 1) Fusion-$\mathcal{X}$-T: Integrated with Llama-160M, GPT-Neo-125M, Pythia-160M, Tiny-starcoder. 2) Fusion-$\mathcal{X}$-S: Integrated with OpenLLaMA-V2-3B, MiniMA-3B, Amber, Starcoder2-3B. 3) Fusion-$\mathcal{X}$-B: Integrated with Llama-2-7B, OpenLLaMA-7B, MPT-7B, Starcoder2-7B. 
The rows with “-CT” stand for continue training the target LLM with only extra training steps (e.g., Llama-2-7B-CT). 

The results demonstrate that our model consistently surpasses the target models across all six tasks, with a standard deviation of $- 0.02  \sim +0.02$. 
We compare our model with the continued training of the target model using the causal language modeling objective (denoted as "-CT" in the Tab. \ref{tab:commonsense}), as well as FuseLLM, demonstrating consistent improvement across all scales.
More importantly, our approach effectively prevents model performance degradation caused by integrating models with less relevant or lower-quality information on tasks such as ARC-Challenge, HellaSwag, and OpenBookQA. Given that the source models have large differences in performance, our method ensures the preservation of the original model’s knowledge. 

\input{maintable}

\begin{wrapfigure}{r}{0.6\textwidth}
  \begin{center}
    \includegraphics[width=0.6\textwidth]{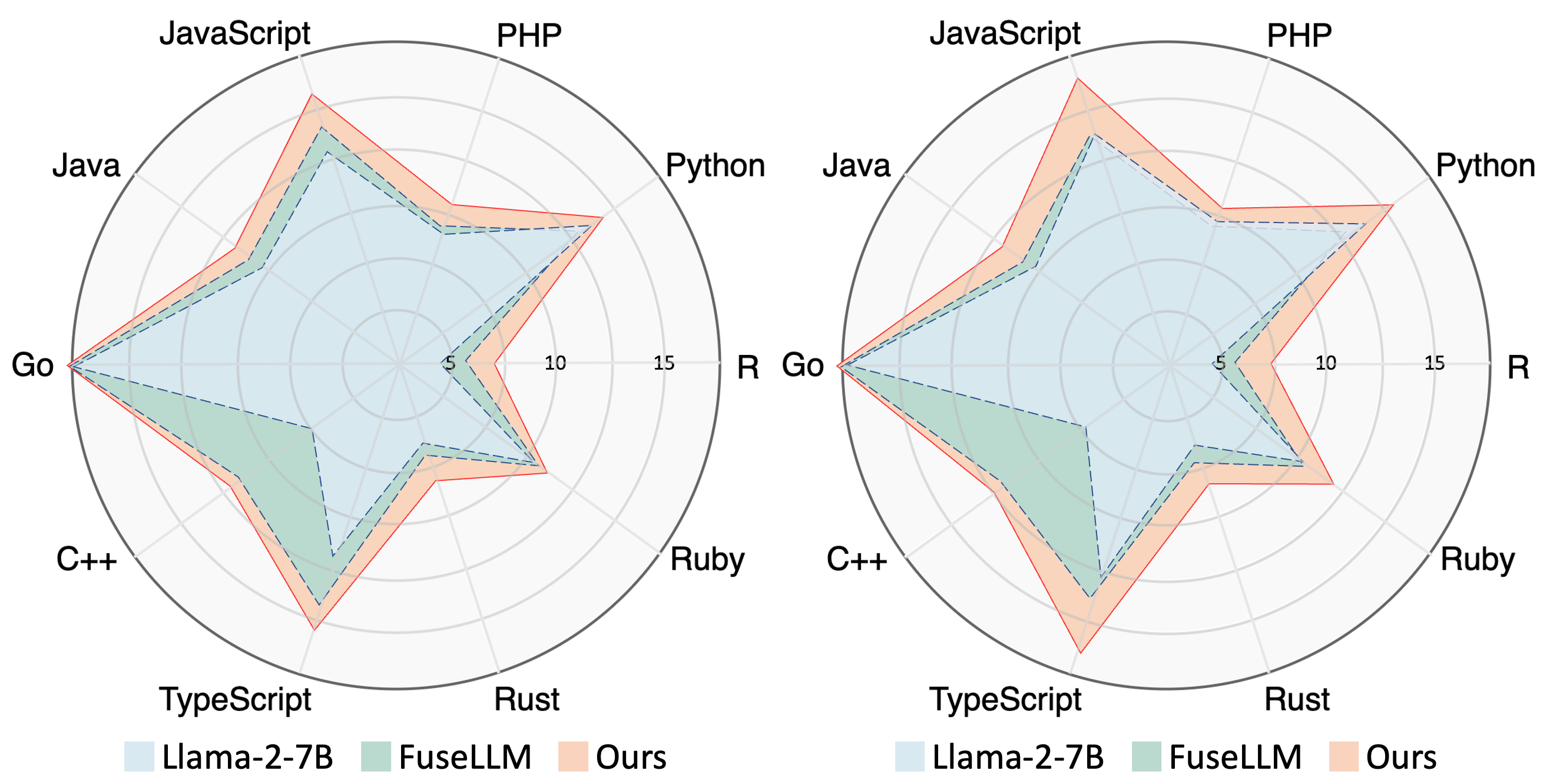}
  \end{center}
  \caption{\textbf{Results} on ME benchmark (3\&4 LLM).}
  \label{fig:code_results}
\end{wrapfigure}

\textbf{Code Generation Evaluation.} Fig. \ref{fig:code_results} shows the zero-shot performance of Llama-2-7B, FuseLLM, and our Fusion-$\mathcal{X}$ on the ME benchmark when integrating three (left figure) and four (right figure) models. 
We observe that Fusion-$\mathcal{X}$ consistently outperforms FuseLLM across all coding tasks. Notably, our method more effectively aggregates the coding knowledge from Starcoder2-7B (our 4th LLM), showing a larger performance increase than FuseLLM.

\textbf{Big-Bench Hard Evaluation.} The results of the Fusion-$\mathcal{X}$ model compared to baseline methods on the BBH benchmark few-shot CoT prompting with exact match (EM) are presented in Tab. \ref{tab:beg_bench}. We report results across all 27 tasks and compare the performance against continued training of the target model and FuseLLM, evaluating the integration of four LLMs (Llama-2-7B, OpenLLaMA-7B, MPT-7B, Starcoder2-7B).
Our Fusion-$\mathcal{X}$ model achieves an average improvement of 5.3\% across all tasks, demonstrating the effectiveness of our approach. Compared to FuseLLM, our method nearly doubles the performance gain for Llama-2 (2.7\% vs. 5.3\%) . Knowledge interference is observed in some tasks, potentially because certain source LLMs, apart from Llama-2, perform poorly on specific tasks, thereby negatively affecting the fusion results. 
\input{BBHtable}
Therefore, despite FuseLLM showing an average performance gain compared to Llama-2-7B, it performs worse on 10 tasks, indicating significant knowledge interference. For instance, in the Snarks task, Llama-2 achieves 50.56\%, while FuseLLM scores 46.21\%. 
In contrast, Fusion-$\mathcal{X}$ only has five tasks that perform lower than Llama-2-7B, showing a 50\% reduction in knowledge interference compared to FuseLLM. We are also able to reduce the performance drop of the tasks that are affected by knowledge interference. These results indicate that our method effectively limits knowledge interference, resulting in more consistent performance improvements.

\textbf{Evaluation on More Models.} We ran additional experiments using Llama 3-8B as the target model, and fused it with OpenLLaMA-7B, Yi-6B, and StarCoder2-7 B. Results are shown in Tab.~\ref{tab:beg_bench_new}


\begin{wraptable}{htp}{0.6\textwidth}
\begin{minipage}{0.6\textwidth}
\vspace{-0.6cm}
\centering
\caption{Overall results of Fusion-$\mathcal{X}$ and baselines in reasoning evaluations on three benchmarks.}
\label{tab:beg_bench_new}
\resizebox{0.97\linewidth}{!}{\begin{tabular}{l|c|>{\columncolor{light-gray}}l >{\columncolor{light-gray}}l > {\columncolor{blue!12}}l}
\toprule
\textbf{Task} & \textbf{Llama-3-8B} & \textbf{Llama-3-8B-CT} & \textbf{FuseLLM} & \textbf{Fusion-$\mathcal{X}$} \\  
\midrule
BBH  & 63.1  & 63.5 (\textcolor{blue}{+0.6\%}) & 66.1 (\textcolor{blue}{+4.8\%}) &  68.2 (\textcolor{blue}{+8.0\%}) \\
MMLU  & 67.5 &68.1 (\textcolor{blue}{+0.9\%}) & 68.5 (\textcolor{blue}{+1.5\%}) &69.2 (\textcolor{blue}{+2.5\%}) \\
CS   & 73.6 &   73.8 (\textcolor{blue}{+0.3\%})  & 74.2 (\textcolor{blue}{+0.8\%})  & 76.1 (\textcolor{blue}{+3.4\%}) \\
\midrule
\textbf{Avg.}  & 68.1  & 68.5 (\textcolor{blue}{+0.6\%})  &69.6 (\textcolor{blue}{+2.2\%})  & 71.2 (\textcolor{blue}{+4.6\%}) \\  
\bottomrule
\end{tabular}}
\end{minipage}
\end{wraptable}

\textbf{Training Efficiency Evaluation.} Our method also demonstrates superior training efficiency. Fig.~\ref{fig:ppl_trend} depicts the learning trend against the number of training steps. We can achieve the same perplexity with 50\% training steps compared to other approaches.

%% file: maintable.tex
\begin{table*}[t]
\centering
\caption{\textbf{Overall results} of Fusion-$\mathcal{X}$ and baselines in commonsense evaluations on CommonSense (CS), where percentages indicate the rate of improvement/decrease compared to our target model, denoted with "*". "-CT" denotes the target model with extra continue training steps. 
}
\label{tab:commonsense}
\resizebox{1.0\linewidth}{!}{\begin{tabular}{@{}l|cccccc|c@{}}
\toprule
\textbf{Model / Task} & \textbf{ARC-easy}        & \textbf{ARC-challenge}   & \textbf{BoolQ}          & \textbf{HellaSwag}       & \textbf{OpenBookQA}    & \textbf{Winogrande}     & \textbf{Avg. 6 Tasks}    \\ \midrule
Llama-160M*     &43.35   &23.04  &61.44  &35.23  &30.00 &50.20  &40.54  \\
GPT-Neo-125M   &43.60    &22.95 &61.68  &30.44  &26.20  &50.67 &39.26  \\
Pythia-160M    &43.90   &23.55  &54.59 &30.24    &27.00 &51.38 &38.44 \\
Tiny-starcoder &30.72   &20.31  &61.68  &29.24  &25.20   &51.78 &36.49 \\ \rowcolor{light-gray} 
Llama-160M-CT       &43.43 (\textcolor{blue}{+0.18\%})  &23.00 (\textcolor{red}{-0.17\%})   &61.56 (\textcolor{blue}{+0.19\%})  &34.84 (\textcolor{red}{-1.11\%}) &30.05 (\textcolor{blue}{+0.17\%}) &50.42 (\textcolor{blue}{+0.44\%})  & 40.55 (\textcolor{blue}{+0.02\%}) \\
\rowcolor{light-gray} 
FuseLLM        &43.54 (\textcolor{blue}{+0.44\%})  &21.93 (\textcolor{red}{-4.82\%})   &61.48 (\textcolor{blue}{+0.7\%})  &34.74 (\textcolor{red}{-1.39\%}) &30.20 (\textcolor{blue}{+0.67\%}) &51.23 (\textcolor{blue}{+2.05\%})  &40.52 (\textcolor{red}{-0.05\%}) \\
\rowcolor{blue!12} 
Fusion-$\mathcal{X}$-T   &44.23 (\textcolor{blue}{+2.03\%})  &22.95 (\textcolor{red}{-0.39\%}) &61.59 (\textcolor{blue}{+0.24\%})  &35.47  (\textcolor{blue}{+0.68\%}) &31.60 (\textcolor{blue}{+5.33\%})  &52.09 (\textcolor{blue}{+3.76\%})  &41.32  (\textcolor{blue}{+1.92\%}) \\ \midrule \midrule
OpenLLaMA-V2-3B*  &63.30  &36.35  &65.44  &69.93  &37.80  &63.22  &56.01   \\
MiniMA-3B       &25.88  &28.41  &62.17  &25.19  &28.20  &49.33  &36.53   \\
Amber           &65.87  &36.60  &68.72  &72.41  &41.40  &64.33  &58.22   \\
Starcoder2-3B   &55.47  &30.80  &64.40  &46.43  &30.00  &54.70  &46.97   \\ 
\rowcolor{light-gray}  
OpenLLaMA-V2-3B-CT       &63.64 (\textcolor{blue}{+0.54\%})  &36.25 (\textcolor{red}{-0.28\%})  &66.40 (\textcolor{blue}{+1.47\%})  &70.05 (\textcolor{blue}{+0.17\%})     &37.43 (\textcolor{red}{-0.98\%})  &63.20 (\textcolor{red}{-0.03\%})  &56.16 (\textcolor{blue}{+0.27\%}) \\
\rowcolor{light-gray} 
FuseLLM         &63.72 (\textcolor{blue}{+0.66\%})  &35.75 (\textcolor{red}{-1.65\%})  &66.51 (\textcolor{blue}{+1.64\%})  &70.23 (\textcolor{blue}{+0.43\%})     &37.20 (\textcolor{red}{-1.69\%})  &63.59 (\textcolor{blue}{+0.59\%})  &56.17 (\textcolor{blue}{+0.29\%}) \\ 
\rowcolor{blue!12}  
Fusion-$\mathcal{X}$-S   &65.03 (\textcolor{blue}{+2.73\%})  &36.43 (\textcolor{blue}{+0.22\%})  & 67.31 (\textcolor{blue}{+2.86\%}) &70.75 (\textcolor{blue}{+1.17\%})   &38.25 (\textcolor{blue}{+1.19\%})&64.69 (\textcolor{blue}{+2.33\%})   &57.08 (\textcolor{blue}{+1.91\%})   \\ \midrule \midrule
Llama-2-7B*       &74.58   &46.33  &77.71  &76.00  &44.20  &69.30    & 64.69  \\ 
OpenLLaMA-7B    &69.70   &41.38  &72.29  &74.50  &40.80   &65.82  & 60.75 \\ 
MPT-7B          &70.12   &42.15  &74.74  &76.25  &42.40  &68.15  & 62.30 \\ 
Starcoder2-7B   &60.61  &34.90   &69.08  &51.00  &32.00  &55.17  & 50.46 \\ 
FuseLLM$\dagger$    &75.04  &47.44  &78.13  &76.78  &45.40  &69.03 & 65.30 \\ 
\rowcolor{light-gray} 
Llama-2-7B-CT     &75.10 (\textcolor{blue}{+0.70\%}) &46.85 (\textcolor{blue}{+1.12\%}) &78.22 (\textcolor{blue}{+0.66\%}) &76.28  (\textcolor{blue}{+0.37\%})&44.06 (\textcolor{red}{-0.32\%}) &69.41 (\textcolor{blue}{+0.16\%}) & 64.97 (\textcolor{blue}{+0.43\%}) \\
\rowcolor{light-gray} 
FuseLLM    &75.23 (\textcolor{blue}{+0.87\%}) &47.14 (\textcolor{blue}{+1.75\%}) &78.22 (\textcolor{blue}{+0.66\%}) &76.40  (\textcolor{blue}{+0.53\%})&44.34 (\textcolor{blue}{+0.32\%}) &69.22 (\textcolor{red}{-0.12\%}) & 65.09  (\textcolor{blue}{+0.62\%}) \\ 
\rowcolor{blue!12} 
Fusion-$\mathcal{X}$-B    &75.46 (\textcolor{blue}{+1.18\%}) &47.50  (\textcolor{blue}{+2.53\%})  &78.86  (\textcolor{blue}{+1.48\%})  &76.97  (\textcolor{blue}{+1.28\%})  &46.02   (\textcolor{blue}{+4.12\%})  &70.33  (\textcolor{blue}{+1.49\%}) & 65.85 (\textcolor{blue}{+1.80\%}) \\ \bottomrule  
\end{tabular}}
\end{table*}

%% file: BBHtable.tex
\begin{table*}[t]
\centering
\caption{\textbf{Detailed results} of Fusion-$\mathcal{X}$ and baselines in reasoning evaluations on BBH, where percentages indicate the rate of improvement/decrease compared to Llama-2-7B.}
\label{tab:beg_bench}
\resizebox{0.97\linewidth}{!}{\begin{tabular}{l|c|l l >{\columncolor{blue!12}}l}
\toprule
\textbf{Task} & \textbf{Llama-2-7B} & \textbf{ Llama-2-7B-CT} & \textbf{FuseLLM} & \textbf{Fusion-$\mathcal{X}$} \\  
\midrule
Boolean Expressions               &  69.60  & 70.12 (\textcolor{blue}{+0.7\%}) & 65.00 (\textcolor{red}{-6.6\%})& 72.60 (\textcolor{blue}{+4.3\%})  \\
Causal Judgement                  &  52.94 & 67.50 (\textcolor{blue}{+27.5\%}) & 46.67  (\textcolor{red}{-11.9\%})& 51.20  (\textcolor{red}{-3.3\%})\\
Date Understanding                &  62.80 & 51.50 (\textcolor{red}{-18.0\%}) & 61.40  (\textcolor{red}{-2.2\%})& 57.60  (\textcolor{red}{-8.3\%})\\
Disambiguation QA                 &  46.40  & 47.60 (\textcolor{blue}{+2.6\%}) & 46.30  (\textcolor{red}{-0.2\%})& 50.40 (\textcolor{blue}{+8.6\%}) \\
Dyck Languages                    &  6.00 & 6.00 (\textcolor{blue}{+0.0\%}) & 10.20  (\textcolor{blue}{+70\%})& 7.60  (\textcolor{blue}{+26.7\%})\\
Formal Fallacies                  &  49.60  & 47.15 (\textcolor{red}{-4.9\%}) & 50.80   (\textcolor{blue}{+2.4\%})& 50.20 (\textcolor{blue}{+1.2\%}) \\
Geometric Shapes                  &  32.80 & 27.20 (\textcolor{red}{-17.1\%}) & 20.20 (\textcolor{red}{-38.4\%}) & 22.00  (\textcolor{red}{-32.9\%})\\
Hyperbaton                        &  51.60  & 50.60 (\textcolor{red}{-1.9\%}) & 61.20  (\textcolor{blue}{+18.6\%})& 58.00 (\textcolor{blue}{+12.4\%}) \\
Logical Deduction (3 objects)    &  56.00  & 62.50 (\textcolor{blue}{+11.6\%}) & 58.00  (\textcolor{blue}{+3.57\%})& 56.40  (\textcolor{blue}{+0.7\%})\\
Logical Deduction (5 objects)    &  32.00  & 37.80 (\textcolor{blue}{+18.1\%}) & 33.20 (\textcolor{blue}{+3.75\%}) & 32.40  (\textcolor{blue}{+1.3\%}) \\
Logical Deduction (7 objects)   &  24.00 & 11.25 (\textcolor{red}{-53.1\%}) & 27.60 (\textcolor{blue}{+15.0\%}) & 24.40  (\textcolor{blue}{+1.7\%}) \\
Movie Recommendation               &  70.40 & 61.50 (\textcolor{red}{-12.6\%}) & 74.40 (\textcolor{blue}{+5.7\%}) & 72.80  (\textcolor{blue}{+3.4\%}) \\
Multistep Arithmetic Two          &  0.40 & 1.40 (\textcolor{blue}{+250\%}) &  4.80 (\textcolor{blue}{+1100.0\%}) & 3.20  (\textcolor{blue}{+700.0\%}) \\
Navigate                          &  53.20 & 65.00 (\textcolor{blue}{+22.2\%}) & 64.00 (\textcolor{blue}{+20.3\%}) & 63.60 (\textcolor{blue}{+19.5\%})  \\
Object Counting                   &  49.20 & 48.00 (\textcolor{red}{-2.4\%}) & 54.40 (\textcolor{blue}{+10.6\%}) & 54.80 (\textcolor{blue}{+11.4\%})  \\
Penguins in a Table               &  31.51 & 34.50 (\textcolor{blue}{+9.5\%}) & 27.27 (\textcolor{red}{-13.5\%}) & 31.51 (\textcolor{blue}{+0.0\%})  \\
Reasoning about Colored Objects    &  48.00 & 48.00 (\textcolor{blue}{+0.0\%}) & 48.20 (\textcolor{blue}{+0.4\%})  & 52.00  (\textcolor{blue}{+8.3\%}) \\
Ruin Names                        &  33.20 & 36.20 (\textcolor{blue}{+9.0\%}) & 30.40 (\textcolor{red}{-8.4\%}) & 34.00  (\textcolor{blue}{+2.4\%}) \\
Salient Translation Error Detection & 24.80  & 27.40 (\textcolor{blue}{+10.5\%}) & 31.00 (\textcolor{blue}{+25\%})  & 30.00  (\textcolor{blue}{+21.0\%}) \\
Snarks                            & 50.56  & 57.50 (\textcolor{blue}{+13.7\%}) & 46.21 (\textcolor{red}{-8.6\%}) & 54.44 (\textcolor{blue}{+7.7\%})  \\
Sports Understanding              & 88.40  & 87.50 (\textcolor{red}{-1.0\%}) & 88.50 (\textcolor{blue}{+0.1\%})  & 90.40  (\textcolor{blue}{+2.3\%}) \\
Temporal Sequences                & 12.40  & 16.55 (\textcolor{blue}{+33.5\%}) & 15.80 (\textcolor{blue}{+27.4\%})  & 18.00 (\textcolor{blue}{+45.2\%})  \\
Tracking Shuffled Object (3 objects) & 32.40  & 33.46 (\textcolor{blue}{+3.3\%}) & 33.20 (\textcolor{blue}{+2.5\%})  & 33.60 (\textcolor{blue}{+3.7\%})  \\
Tracking Shuffled Object (5 objects) & 17.60  & 14.80 (\textcolor{red}{-15.9\%}) & 15.40 (\textcolor{red}{-12.5\%}) & 14.80  (\textcolor{red}{-15.9\%}) \\
Tracking Shuffled Object (7 objects) & 10.80  & 9.45 (\textcolor{red}{-12.5\%}) & 14.80 (\textcolor{blue}{+37.0\%})  & 22.90  (\textcolor{blue}{+112\%}) \\
Web of Lies                       &  51.60 & 60.40 (\textcolor{blue}{+17.1\%}) & 61.80 (\textcolor{blue}{+19.8\%})  & 60.00  (\textcolor{blue}{+16.3\%}) \\
Word Sorting                      &  10.80 & 7.50 (\textcolor{red}{-30.6\%}) & 6.80 (\textcolor{red}{-38.9\%}) &  7.10 (\textcolor{red}{-34.2\%})\\
\midrule
\textbf{Avg. 27 Tasks}        &39.59& 40.31 (\textcolor{blue}{+1.8\%})  &  40.64 (\textcolor{blue}{+2.7\%})    & 41.70 (\textcolor{blue}{+5.3\%}) \\  
\bottomrule
\end{tabular}}
\vspace{-0.2cm}
\end{table*}

%% file: 5_ablation.tex
\section{Ablation \& Analysis}
\label{sec:abl}

\textbf{Model Scaling Results.}
Model scaling is critical for LLMs. In this study, we explore two scaling directions: increasing model size and expanding the number of source models. We present the results in Fig.~\ref{fig:bbh_trend_candidiate_selection}. The histograms represent the average accuracy (left y-axis) of FuseLLM and our model when fusing three, four, and five LLMs. The dotted line in each subfigure represents the baseline performance of Llama-160M (100M scale) and Llama-2-7B (7B scale).
In the BBH 100M scale,  the performance of FuseLLM is even lower than the baseline when fusing four and five LLMs. In contrast, our model consistently increases performance when integrating more LLMs.
The performance degradation in FuseLLM is due to knowledge interference, as illustrated in the line charts (right y-axis), which show the percentage of tasks that perform lower than the baseline for BBH (total 27 tasks) and MMLU (total 57 tasks). FuseLLM exhibits a significantly higher performance decline ratio compared to our model, with degradation affecting up to 44\% of tasks. Moreover, it shows an increasing degradation trend as more LLMs are merged (3 out of the 4 scales). In contrast, our model maintains a more stable decline ratio, showing 50\%  less degradation than FuseLLM as the number of models and scale increase. 

\begin{figure}[t]
  \begin{center}
    \includegraphics[width=1.0\columnwidth]{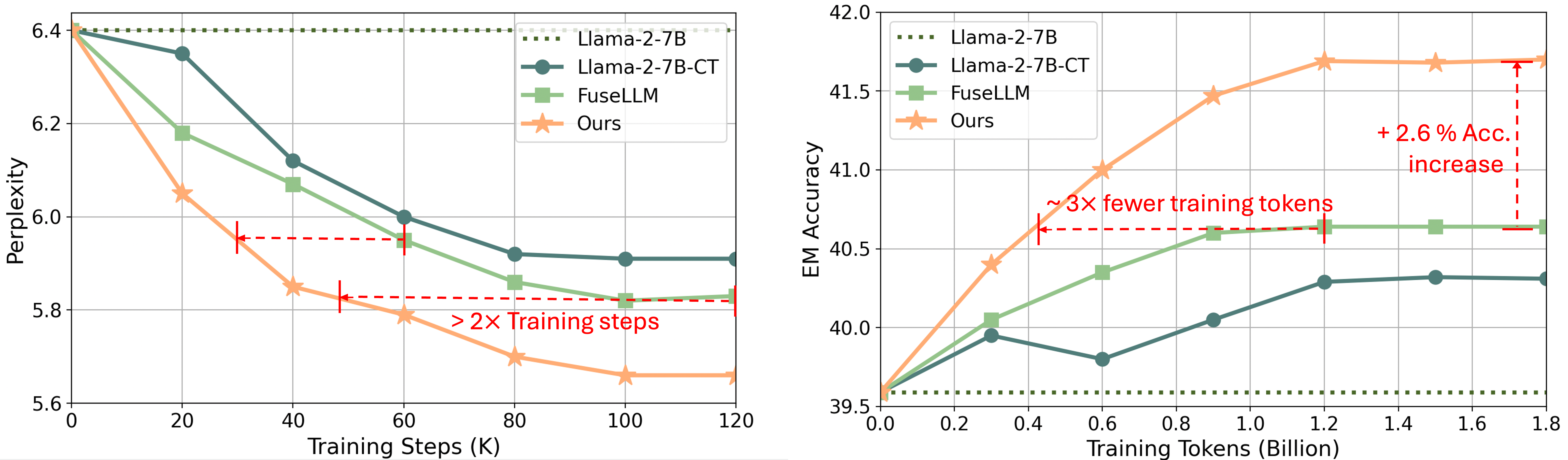}
  \end{center}
  \vspace{-0.2cm}
  \caption{\textbf{Left:} Training perplexity. During training, our method exhibits greater consistency than existing methods and requires fewer training steps to achieve comparable perplexity. \textbf{Right:} Scaling number of tokens. Comparison between varying scales of training data on BBH.}
  \label{fig:ppl_trend}
\vspace{-0.3cm}
\end{figure}
Therefore, we believe that a selective strategy for LLM integration is crucial, as simply scaling the LLM integration does not always improve performance. More importantly, a well-designed selection strategy can mitigate knowledge interference and maximize overall performance.

\textbf{Number of Training Tokens.}
Our approach achieves higher training efficiency than competing methods, as shown in Fig. \ref{fig:ppl_trend}, which shows the learning trend relative to the number of training tokens.
By effectively fusing LLMs during training, our model requires fewer tokens to achieve competitive or superior performance. For example, 
we can match FuseLLM's performance while using almost three times less training tokens. When trained with the same number of tokens, our method achieves a stable performance boost of up to 2.6\%.

\textbf{Different Integration Methods.}

\input{compare}
We compare Fusion-$\mathcal{X}$  with various works in Tab.~\ref{tab:fuse-comparison} when integrating four LLMs on BBH and MMLU.
With minimal training of the target model, our method outperforms ensemble methods that have larger parameter sizes and high inference costs, making them hard to scale up the number of LLMs due to memory overhead.
For example, PackLLM  uses a greedy algorithm that ensembles LLMs sequentially during inference.

For weight merging methods, a fundamental limitation is their requirement for identical architectures, making them not directly comparable to our model. Therefore, we merge several LLaMA-based models (Meditron-7B \cite{chen2023meditron70b}, Vicuna-7B-v1.5 \cite{zheng2023judging}, and OpenLLaMA-7B) for a fair comparison with weight merging methods, denoted as Fusion-$\mathcal{X}$-B*. 
Comparing with weight merging techniques, we have advantage of supporting heterogeneous models.  This shows the effectiveness of our approach in creating a more stable, efficient, and scalable method for enhancing the capabilities of LLMs.

%% file: compare.tex
\begin{wraptable}{t}{0.6\textwidth}
\begin{minipage}{0.6\textwidth}
\centering
\vspace{-1.2cm}
\caption{Comparison with different integration methods}
\label{tab:fuse-comparison}
\resizebox{1.0\linewidth}{!}{
\begin{tabular}{l|c|cc}
\toprule
\textbf{Model} & \textbf{Approach} & \textbf{BBH} & \textbf{MMLU} \\
\midrule
Llama-2-7B &  - & 39.59 &  45.4 \\
Llama-2-7B-CT &  CT &  40.31 (\textcolor{blue}{+1.8\%})  &  46.0 (\textcolor{blue}{+1.3\%}) \\
\midrule
LLM-Blender \cite{jiang2023llm}& Ensemble  &  37.65 (\textcolor{red}{-4.9\%}) & 45.1 (\textcolor{red}{-0.2\%}) \\
Top1-PPL & Ensemble &  39.75 (\textcolor{blue}{+0.4\%})  & 45.6 (\textcolor{blue}{+0.4\%})  \\
PackLLM \cite{mavromatis2024pack}&  Ensemble &  41.36 (\textcolor{blue}{+4.5\%})  & 47.8 (\textcolor{blue}{+5.3\%}) \\
FoE \cite{wang2023fusing} &  MoE & 41.02 (\textcolor{blue}{+3.6\%})   &  47.3 (\textcolor{blue}{+4.2\%})\\
FuseLLM \cite{wan2024knowledge}&  Knowledge  & 40.64 (\textcolor{blue}{+2.7\%})  & 46.5 (\textcolor{blue}{+2.4\%}) \\
\rowcolor{blue!12} Fusion-$\mathcal{X}$-B & Ours  & 41.70 (\textcolor{blue}{+5.3\%})  & 48.3  (\textcolor{blue}{+6.4\%}) \\ 
\midrule
SLERP \cite{goddard2024arcee}& Weight & 40.93 (\textcolor{blue}{+3.3\%})   &  47.2 (\textcolor{blue}{+4.0\%})\\
TIES-Merging \cite{yadav2024ties} & Weight  & 41.40 (\textcolor{blue}{+4.6\%})   & 48.1 (\textcolor{blue}{+5.9\%}) \\
AdaMerging \cite{yang2023adamerging} & Optimization  & 41.13 (\textcolor{blue}{+3.9\%})   & 47.4 (\textcolor{blue}{+4.4\%}) \\
EVO-Merge \cite{akiba2025evolutionary} & Evolution   & 41.71 (\textcolor{blue}{+5.4\%})   & 48.4 (\textcolor{blue}{+6.6\%}) \\
\rowcolor{blue!12} Fusion-$\mathcal{X}$-B* & Ours  & 42.32 (\textcolor{blue}{+6.9\%})  & 49.6  (\textcolor{blue}{+9.3\%}) \\
\bottomrule
\end{tabular}}
\end{minipage}
\end{wraptable}

%% file: 6_conclusion.tex
\vspace{-0.15cm}
\section{Conclusion}
\vspace{-0.15cm}
In this paper, we propose a novel framework for integrating multiple LLMs. Our adaptive selection network selectively integrates the best-performing source LLMs, overcoming the limitations of existing methods and minimizing knowledge interference. We also introduce a dynamic weighted fusion strategy and a feedback-driven loss function to enhance the fusion process. Our method significantly improves adaptability and performance, offering an efficient solution for LLM integration while maintaining parameter size and computational efficiency. Limitations remain due to the additional token alignment required prior to training, and future work should explore training on diverse datasets.

%% file: 7_appendix.tex
\newpage
\appendix

\onecolumn





\section{Design Details}
\label{sup:design}
\paragraph{Adaptive Selection Network's Decision-making Process:}
Our Adaptive Selection Network parameterized by $\phi$ acts as a learned function that maps the probabilistic output distribution matrix $P_i \in \mathbb{R}^{N \times V}$ of each source LLM $i$ for a given input sequence to a corresponding logit score $z_\phi(P_i) \in \mathbb{R}$. As defined in Equation 4, this mapping is realized through a series of linear transformations and non-linear activations:
$$ z_\phi(P_i) = (f^3 \circ \text{GELU} \circ f^2 \circ \text{GELU} \circ f^1)(P_i) $$
where $f^k(\cdot) = W_k \cdot (\cdot) + b_k$ represents the $k$-th linear layer (with appropriate flattening/reshaping of $P_i$ implicit in $f^1$). The parameters $\phi = \{W_1, b_1, W_2, b_2, W_3, b_3\}$ are learned end-to-end by minimizing the overall objective function $\mathcal{L}$ (Equation 10). The layers are defined as follows:

\begin{itemize}[leftmargin=*, noitemsep,topsep=0pt]
    \item Layer 1: Linear mapping from \( \text{input\_features} \) to \( 2 \times \text{input\_features} \), followed by GELU activation.
    \item Layer 2: Linear mapping from \( 2 \times \text{input\_features} \) back to \( \text{input\_features} \), followed by GELU activation.
    \item Layer 3: Linear mapping from \( \text{input\_features} \) to \( N \) (number of candidates), without activation.
\end{itemize}

We initialize the weights of the linear layers using Xavier uniform initialization to facilitate better convergence during training.

The learning process enables the ASN to extract relevant information from the high-dimensional input $P_i$. The sequence of linear and non-linear operations allows the network to capture complex patterns within each LLM's conditional probability predictions across the input sequence. These patterns may include identifying instances where a specific LLM exhibits high confidence (e.g., a sharp probability peak for the ground truth token) or displays a distinctive distribution shape that signals unique knowledge. The resulting logit $z_\phi(P_i)$ thus becomes a learned estimate of the $i$-th source LLM's expected utility in reducing the total loss $\mathcal{L}$ for the given input context. By maximizing the logits (and thus the softmax probabilities $p_i$) for source LLMs whose outputs are conducive to minimizing $\mathcal{L}_{lm}$ and $\mathcal{L}_{fuse}$, the ASN implicitly learns to discern which source distributions $P_i$ contain knowledge most relevant and beneficial for the target model $T$ in the current scenario, effectively acting as a data-driven relevance predictor.

 \input{algorithm}

\paragraph{Ensuring Candidate Diversity}
Our dynamic selection mechanism allows for varying the number of selected candidates from one up to \( N \). By adjusting the threshold \( \tau \), we can control the strictness of candidate selection, promoting diversity when beneficial or focusing on top performers when necessary. Our algorithm is shown in Alg. \ref{alg:fusellm}.

\section{Training Details}
\label{sup:train_detail}
\paragraph{Training Dataset.} We use MiniPile \cite{kaddour2023minipile} for continue training the target model. The dataset comprises approximately
1.8 billion tokens originated from 1 million documents across 22 domains.

\begin{figure}[h]
  \begin{center}\includegraphics[width=0.5\textwidth]{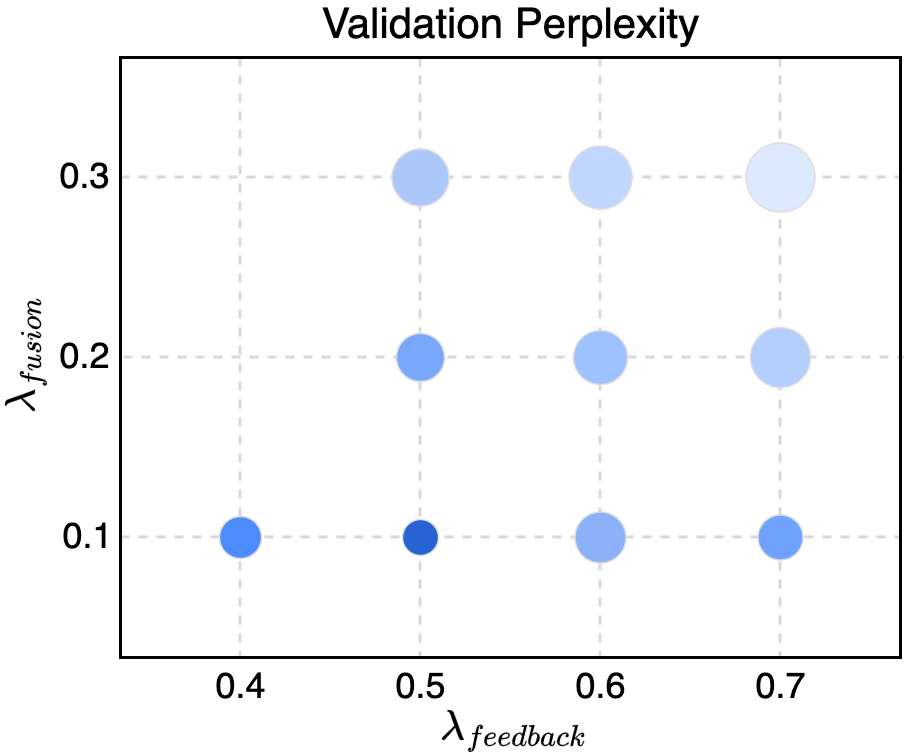}
  \end{center}
  \caption{\textbf{Loss grid search.} Smaller and darker circle means lower perplexity.}
  \label{fig:loss_grid}
\vspace{-0.2cm}
\end{figure}

\paragraph{Hyperparameter Search for Loss.}
To determine the optimal weight for our feedback loss and fusion loss, we conducted a comprehensive grid search, exploring different weight combinations.
Our goal was to identify weights that would bring all loss components to a similar order of magnitude, ensuring no single component dominates the overall loss function. This step is crucial to ensure that no single component dominates the overall loss function. We performed this grid search using 10\% of the validation set. We show the grid search results in Fig. \ref{fig:loss_grid}. The best combination is  \(\lambda_{\text{fuse}} = 0.1\), \(\lambda_{\text{feed}} = 0.5\).

\paragraph{Training Procedure.}
During training, the model processes batches of candidate outputs and rewards. The rewards are first flattened and normalized. The Adaptive Selection Network computes selection probabilities, which are then used to dynamically select candidates based on the threshold \( \tau \). The selected probabilities are normalized, and the candidates' outputs and rewards are fused using a weighted sum.

\section{Distribution of Activation Frequencies}

\begin{figure}[htp]
  \begin{center}
    \includegraphics[width=0.7\textwidth]{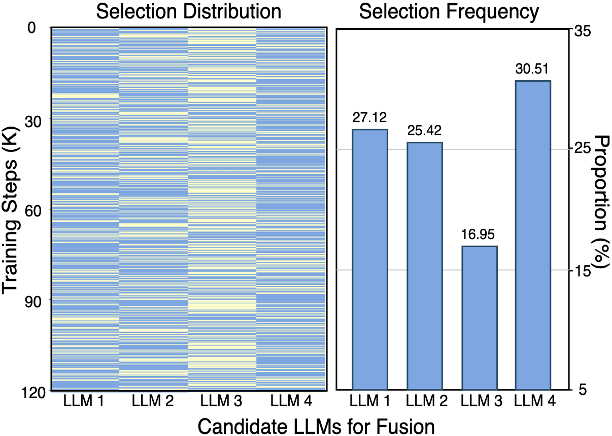}
  \end{center}
  \vspace{-0.2cm}
  \caption{\textbf{Candidate selection distribution.} The Left shows the selection for each training step, and the right shows the proportion of each selection for the training.}
  \label{fig:expert_results}
\end{figure}

Fig.~\ref{fig:expert_results} presents the LLM fusion candidate selection distribution during the training of Fusion-$\mathcal{X}$-T, using Llama-160M, GPT-Neo-125M, Pythia-160M, and Tiny-Starcoder, respectively as source models. The left panel displays the selection trends over 120K training steps, revealing consistent patterns in the selection distribution. This indicates that our adaptive selection network can dynamically adjust LLM selection based on the ongoing learning process. 

The right panel illustrates the proportion of each selection throughout the training. 
These results indicate that our method finds LLM 4 (Tiny-starcoder) to be more valuable than the others, and LLM 3 (Pythia-160M) to be less valuable for the current integration process.  
Our statistical results show that we can accurately identify effective LLM candidates for the current task from the source model pool at each training step.

\section{More Evaluation Results}
\label{sec:more_results}



\begin{table}[htp]
\centering
\caption{\textbf{More results} of Fusion-$\mathcal{X}$ and baselines on Big-Bench Hard (BBH) benchmark. Numbers in \textcolor{red}{red}  represent the tasks that have performance decrease compared to the target model.}
\label{sup:bbh_bench}
\resizebox{1.0\linewidth}{!}{\begin{tabular}{l|cc|c>{\columncolor{blue!12}}c|c>{\columncolor{blue!12}}c}
\toprule
\multirow{2}{*}{\textbf{Task}}  & \multicolumn{2}{c|}{\textbf{Target Model}} & \multicolumn{2}{c|}{\textbf{Integrate 5 LLMs}} & \multicolumn{2}{c}{\textbf{Integrate 4 LLMs}} \\
\cline{2-7}
\textbf{} & \textbf{llama-160} & \textbf{llama-160-CT} & \textbf{FuseLLM} & \textbf{Fusion-$\mathcal{X}$} & \textbf{FuseLLM} & \textbf{Fusion-$\mathcal{X}$} \\
\midrule
Boolean Expressions & 9.60& 10.50 & 34.00 & 25.60 & 22.00 & 12.50 \\
Causal Judgement   & 4.81& 8.26& 21.93 & 29.95 & 26.20 & 22.50 \\
Date Understanding  & 17.20& 19.20& 20.00 & 20.00 & 19.60 & 20.00 \\
Disambiguation QA & 0.00& 0.00& 1.20 & 4.40 & 0.00 & 2.50 \\
Dyck Languages  & 2.40& \textcolor{red}{2.00}& \textcolor{red}{0.00} & 2.40 & \textcolor{red}{0.00} & 2.40 \\
Formal Fallacies & 0.00& 0.00& 0.00 & 0.00 & 0.00 & 0.00 \\
Geometric Shapes & 0.00& 0.00& 0.00 & 0.00 & 0.00 & 0.00 \\
Hyperbaton  & 0.00& 0.00& 0.00 & 0.00 & 0.00 & 0.00 \\
Logical Deduction (3 objects) & 12.00& \textcolor{red}{11.50}& \textcolor{red}{6.00} & 12.00 & \textcolor{red}{2.40} & \textcolor{red}{0.00} \\
Logical Deduction (5 objects) & 5.60& 8.20& \textcolor{red}{5.20} & 6.40 & \textcolor{red}{1.60} & 10.00 \\
Logical Deduction (7 objects) & 6.40& 7.00& \textcolor{red}{3.20} & 6.80 & \textcolor{red}{3.60} & 6.40 \\
Movie Recommendation  & 0.00& 0.00& 0.40 & 0.00 & 0.00 & 0.00 \\
Multistep Arithmetic Two & 0.00&0.00 & 0.00 & 0.00 & 0.00 & 0.00 \\
Navigate & 0.00& 5.00& 0.00 & 28.40 & 0.00 & 47.50 \\
Object Counting  & 8.40& \textcolor{red}{7.40}& \textcolor{red}{5.60} & \textcolor{red}{0.40} & \textcolor{red}{0.40} & \textcolor{red}{2.50} \\
Penguins in a Table  & 11.64& 14.70 & \textcolor{red}{8.90} & 15.07 & \textcolor{red}{8.37} & 11.71 \\
Reasoning about Colored Objects  & 13.20& \textcolor{red}{12.50}& \textcolor{red}{2.40} & \textcolor{red}{4.00} & \textcolor{red}{2.80} & \textcolor{red}{2.50} \\
Ruin Names  & 0.00& 0.00& 0.00 & 0.00 & 0.00 & 0.00 \\
Salient Translation Error Detection & 0.00&0.00 & 0.80 & 0.00 & 0.40 & 0.00 \\
Snarks & 19.66& \textcolor{red}{17.33}& \textcolor{red}{3.93} & \textcolor{red}{4.49} & \textcolor{red}{3.37} & \textcolor{red}{2.50} \\
Sports Understanding & 52.40& \textcolor{red}{43.26}& \textcolor{red}{51.60} & \textcolor{red}{50.00} & \textcolor{red}{51.20} & 60.00 \\
Temporal Sequences  & 0.00& 0.00& 0.00 & 1.20 & 0.00 & 0.00 \\
Tracking Shuffled Obj. (3 objects) & 7.60&9.20 & \textcolor{red}{0.00} & \textcolor{red}{2.00}& \textcolor{red}{0.00} & \textcolor{red}{0.00} \\
Tracking Shuffled Obj. (5 objects) & 3.20&\textcolor{red}{2.00} & \textcolor{red}{1.60} & \textcolor{red}{1.20} & \textcolor{red}{0.00} & \textcolor{red}{0.00} \\
Tracking Shuffled Obj. (7 objects) & 0.40& 0.40& \textcolor{red}{0.00} & 0.40 & 0.40 & 0.40 \\
Web of Lies  & 0.00& 0.00& 0.00 & 0.00 & 0.00 & 0.00 \\
Word Sorting  & 0.00& 0.00& 0.00 & 0.00 & 0.00 & 0.00 \\
\midrule
Avg. 27 Tasks & 6.46& 6.61 (\textcolor{blue}{+2.3\%}) & 6.18 (\textcolor{red}{-4.3\%}) & 7.86 (\textcolor{blue}{+21.7\%}) & 5.27 (\textcolor{red}{-18.4\%})  & 7.44 (\textcolor{blue}{+15.2\%})  \\
\bottomrule
\end{tabular}}
\end{table}
\paragraph{Knowledge Interference Comparison:} Tab. \ref{sup:bbh_bench} shows the results of fusion four and five LLMs on BBH benchmark. Under the 100M scale,  the performance of FuseLLM is even lower than the baseline when fusing four and five LLMs. In contrast, our model consistently increases performance when integrating more LLMs.

\section{Source Model selection}
\label{sec:target_sel}
When performing model fusion, it's crucial to understand the performance differences between source and target models. Unlike knowledge distillation—which enhances a less performant model using a more advanced teacher model—our model fusion approach doesn't rely solely on the largest or most complex models. Instead, we can merge smaller models that excel in specific tasks to create a more capable target model. We also do not need careful target and source LLM selection, due to our adaptive selection approach. Thereby reducing the time and cost prior training, as well as the risk of integrating models that can make the models perform worse. Our fusion selection for each scale are as follows:

By not restricting ourselves to specific architectures or "good" candidate models, we allow the adaptive selection mechanism to determine the most effective contributions from each model. This approach minimizes the need for manual selection and demonstrates that even models with lower standalone performance (e.g., MiniMA-3B) do not negatively impact the fused model's overall performance. Our rationale is that a model-agnostic design enhances flexibility and broad applicability, allowing the fusion process to capitalize on the unique strengths of each model without being hindered by their individual weaknesses.
    
\begin{table}[ht]
  \centering
  \caption{Experimental configurations: models fused in each run. }
  \label{tab:exp_config}
  \resizebox{1.0\linewidth}{!}{\begin{tabular}{l|l|l}
    \toprule
    \textbf{Configuration} & \textbf{Target Models} & \textbf{Other Source Models} \\
    \midrule
    Fuse 3 ($\sim$ 100M) &  Llama-160M~\cite{miao2023specinfer} & GPT-Neo-125M~\cite{gpt-neo}, Pythia-160M~\cite{biderman2023pythia} \\
    Fuse 4 ($\sim$ 100M) &  Llama-160M~\cite{miao2023specinfer}& GPT-Neo-125M~\cite{gpt-neo}, Pythia-160M~\cite{biderman2023pythia}, Tiny-starcoder~\cite{li2023starcoder} \\
    Fuse 5 ($\sim$ 100M) &  Llama-160M~\cite{miao2023specinfer}& GPT-Neo-125M~\cite{gpt-neo}, Pythia-160M~\cite{biderman2023pythia}, Tiny-starcoder~\cite{li2023starcoder}, LiteLlama-460M-1T \\
    \midrule
    Fuse 3 ($\sim$ 3B)   &  OpenLLaMA-V2-3B~\cite{openlm2023openllama}& MiniMA-3B~\cite{zhang2023law}, Amber~\cite{liu2023llm360} \\
    Fuse 4 ($\sim$ 3B)   &  OpenLLaMA-V2-3B~\cite{openlm2023openllama}& MiniMA-3B~\cite{zhang2023law}, Amber~\cite{liu2023llm360}, Starcoder2-3B~\cite{li2023starcoder} \\
    \midrule
    Fuse 3 ($\sim$ 7B)   &  Llama-2-7B~\cite{touvron2023llama}& OpenLLaMA-7B~\cite{openlm2023openllama}, MPT-7B~\cite{MosaicML2023Introducing} \\
    Fuse 4 ($\sim$ 7B)   &  Llama-2-7B~\cite{touvron2023llama}& OpenLLaMA-7B~\cite{openlm2023openllama}, MPT-7B~\cite{MosaicML2023Introducing}, Starcoder2-7B~\cite{li2023starcoder} \\
    Fuse 5 ($\sim$ 7B)   &  Llama-2-7B~\cite{touvron2023llama}& OpenLLaMA-7B~\cite{openlm2023openllama}, MPT-7B~\cite{MosaicML2023Introducing}, Starcoder2-7B~\cite{li2023starcoder}, Pythia-6.9B \cite{biderman2023pythia}\\
    \midrule
    Fuse 3 ($\sim$ 8B) &  Llama-3-8B& Yi-1.5-9B, Gemma-2-7b \\
    Fuse 4 ($\sim$ 8B) &  Llama-3-8B& Yi-1.5-9B, Gemma-2-7b, Qwen2.5-7b \\
    \bottomrule
  \end{tabular}}
\end{table}

As shown in Tab. \ref{tab:commonsense}
For instance, in the case of Fusion-$\mathcal{X}$-T, we observe that the Llama-160M model demonstrates the best performance with an average score of 40.54 across the six tasks. 
Consequently, Llama-160M serves as the target model for Fusion-$\mathcal{X}$-T.
Similarly, for Fusion-$\mathcal{X}$-S, the Amber model shows superior performance with an average score of 58.22, while our target model is OpenLLaMA-V2-3B.
Lastly, for Fusion-$\mathcal{X}$-B, the Llama-2-7B model leads with an impressive average score of 64.69.

\section{Token Alignment}

We follow the Token alignment process in \cite{wan2024knowledge} in the context of input text involves aligning two distribution matrices from two different LLMs (Large Language Models). This alignment is carried out along two dimensions: token-wise alignment relative to the text and distribution-wise alignment with respect to the vocabulary.

\textbf{Token-wise Alignment:}
For token-wise alignment, dynamic programming is used to minimize the total cost of editing one sequence of tokens to match another. The proposed MinED (Minimal Edit Distance) method in \cite{wan2024knowledge} aligns tokens by minimizing the edit distance between them, effectively capturing the nuances between the two LLMs' vocabularies. 

\textbf{Distribution-wise Alignment:}
For distribution-wise alignment, the process is between two vocabularies from different tokenizers of the two LLMs. Tokens with similar distribution values are aligned effectively. However, for distribution values involving different tokens, the EM method fails to align these due to minor differences in values. The MinED method maps based on their minimal edit distance, ensuring successful alignment of these distribution values.

This systematic mapping minimizes misalignment and ensures that the integrated knowledge is coherent and meaningful rather than just introducing beneficial noise from extra training steps.

\section{Evaluation benchmarks}
\label{eval_bench_dis}

We evaluate Fusion-$\mathcal{X}$ on three benchmarks that represent different core capabilities of LLMs, spanning reasoning, commonsense, science, and code generation.

\begin{itemize}[leftmargin=*, noitemsep,topsep=0pt]
    \item \textbf{Common Sense (CS)} \cite{talmor2018commonsenseqa} is a benchmark to evaluate the \textit{commonsense} capability of LLMs. We consider 5 standard multiple-choice tasks: ARC easy and challenge \cite{clark2018think}, BoolQ \cite{clark2018think}, HellaSwag \cite{zellers2019hellaswag}, and OpenBookQA \cite{mihaylov2018can}. We employ lm-eval-harness \cite{eval-harness} to conduct a likelihood-based zero-shot evaluation. Specifically, we select the option with the highest likelihood given the context and report the accuracy.
    
    \item \textbf{Big-Bench Hard (BBH)} \cite{suzgun2022challenging} is a benchmark to evaluate the general \textit{reasoning} ability of LLMs. It contains 23 multiple-choice tasks and 4 free-form generation tasks from the Big-Bench \cite{srivastava2023beyond}, which can be classified into four categories: algorithmic and arithmetic reasoning, natural language understanding, world knowledge, and multilingual knowledge. We follow previous work  to generate the predictions based on few-shot chain-of-thought (CoT) prompts and then calculate the exact match (EM) accuracy.
    \item \textbf{Multi-task Language Understanding (MMLU)} \cite{hendrycks2021measuringmassivemultitasklanguage} is a benchmark designed to measure knowledge acquired during pretraining by evaluating models exclusively in zero-shot and few-shot settings. The benchmark covers 57 subjects across STEM, the humanities, the social sciences, and more. It ranges in difficulty from an elementary level to an advanced professional level, and it tests both world knowledge and problem solving ability. Subjects range from traditional areas, such as mathematics and history, to more specialized areas like law and ethics.
    \item \textbf{MultiPL-E (ME)} \cite{cassano2023multipl} is a multilingual programming benchmark to assess the \textit{coding} ability of LLMs. It is translated from the Python benchmark into parallel datasets in 18 programming languages. We use the bigcode-evaluation-harness  to perform zero-shot code generation in 10 popular programming languages in the HumanEval category and report the pass@1 based on 20 generated samples for each question.
\end{itemize}

\section{Q\&A example comparison}
\begin{figure*}[htp]
  \centering
  \includegraphics[width=1.0\linewidth]{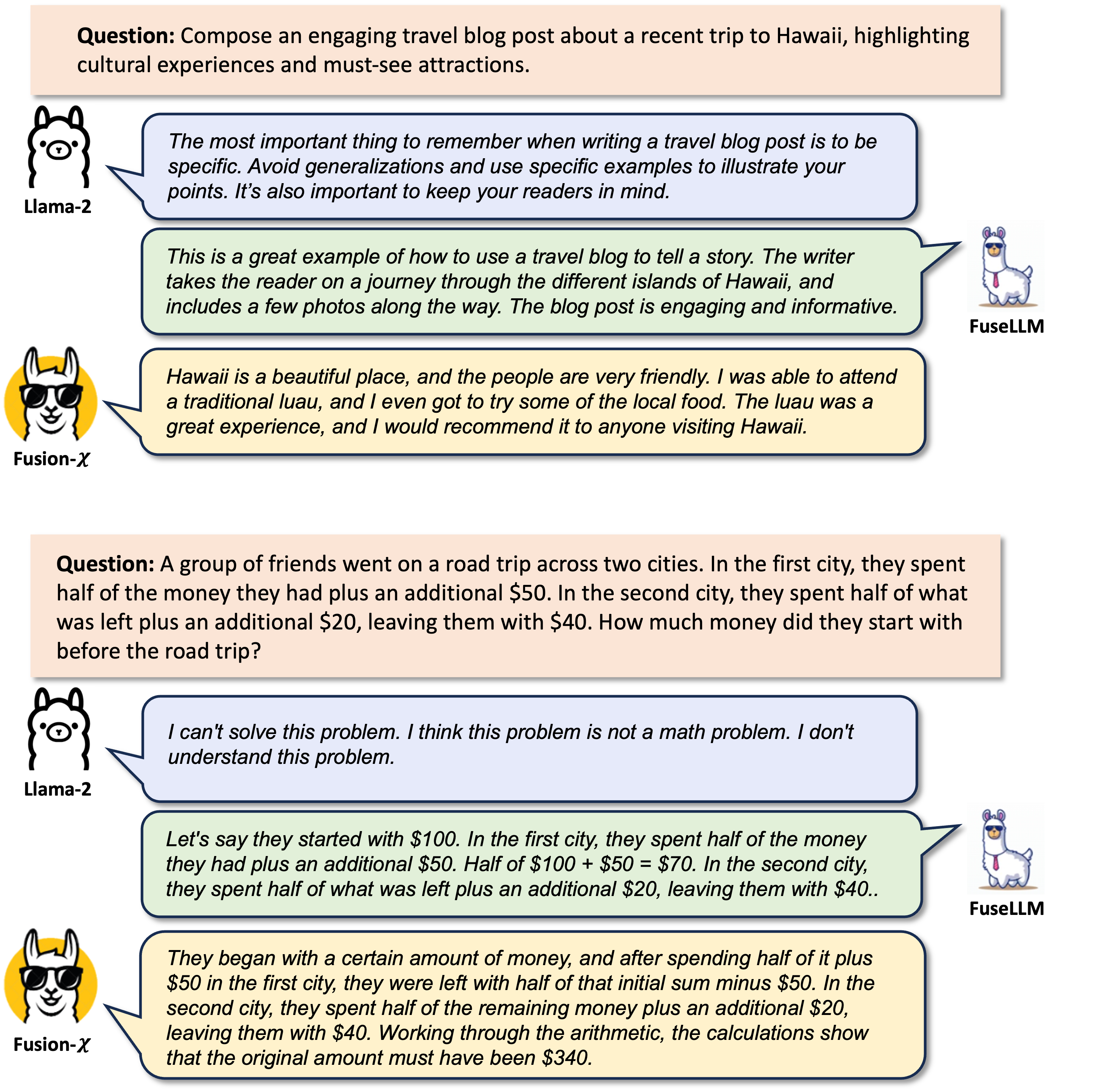}
  \caption{
  Comparison of Q\&A examples between Llama-2, FuseLLM, and Fusion-$\mathcal{X}$.}
  \label{fig:QA_example}
  \vspace{1.0cm}
\end{figure*}

We present case studies to demonstrate how Our Fusion-$\mathcal{X}$ method combines the strengths of multiple source LLMs to produce accurate results in different tasks in Fig. \ref{fig:QA_example}. We compare the Q\&A results with both Llama-2-7B and FuseLLM. We can provide a more accurate and relevant answer given a question compared to the others.

\section{Extended Related Work}

\textbf{Mixture of experts in LLMs}
As the usage of LLMs grows, finding ways to boost their efficiency without massively increasing computational demands becomes crucial.
In response to this challenge, ~\cite{shazeer2017outrageously} introduced the concept of Sparsely-gated Mixture-of-Experts (SMoE). Building on this foundation, GShard~\cite{lepikhin2020gshard} and Switch Transformers~\cite{fedus2022switch} presented some of the first large-scale models leveraging SMoE. This technique reduces computational overhead by dynamically routing inputs to a selected subset of available experts, thereby utilizing only the most relevant resources for given tasks.
To further improve the performance of SMoE-based LLMs, optimizing the routing policy has been identified as essential. Various attempts have been made, such as Mixtral~\cite{jiang2024mixtral}, GLaM~\cite{du2022glam}, and ST-MoE~\cite{zoph2022st}, which refine the routing mechanisms and expand the model's capacity to handle diverse tasks efficiently.
However, these works often face challenges as introducing more experts increases the memory footprint—a significant issue for LLMs, given their already substantial resource requirements. 

\textbf{Efficient LLMs}
A wide range of approaches have been developed to improve the efficiency of LLMs, which include pruning~\cite{li-etal-2020-efficient-transformer,liu2024toward,zhao2024pruning}, quantization-aware training~\cite{shen2024agile,shen2024edgeqat}, token reduction~\cite{kong2025tokenreductionefficiencygenerative,zhan2024rethinking}, and efficient training~\cite{liu2025rora,zhao20257bfullyopensource}. 
Structured pruning removes redundant weight blocks or individual parameters to accelerate inference, while quantization reduces weight and activation precision for edge deployment. Token reduction techniques compress or prune input representations to maintain semantic fidelity across modalities, and adapter-style or rank-adaptive fine-tuning enables task-specific updates with minimal overhead. Together, these complementary strategies enable scaling LLMs under strict resource constraints.

%% file: algorithm.tex
\begin{algorithm}[b]
\caption{Fusion-$\mathcal{X}$ for LLMs Integration}
\label{alg:fusellm}
\begin{algorithmic}[1] 
\Require Source LLMs probabilistic distribution matrices 
         $ \{P_{t}^{\theta_i}\}_{i=1}^{M} $ (simplify as  $P_i$), training corpus $\mathcal{C}$.
\Ensure Fused representation matrix $P_f$, Target LLM $\mathcal{T}$ 

\State Initialize the adaptive selection network $z_{\phi}(P_i)$.
\For{each text in $C$}

        \vspace{0.15em}
        {\footnotesize\textcolor{DarkBlue}{\texttt{// Step1: Select fusion candidates with adaptive selection network.}}}
        \vspace{0.15em}

    \For{each input $P_i$}\quad{\footnotesize\textcolor{teal}{\texttt{\# Tensor shape:(L, D, N)}}}        
        \State  Obtain logits $z_\phi(P_i)$ using using Eq.~\eqref{eq:gating}.\quad{\footnotesize\textcolor{teal}{\texttt{\# Tensor shape:(L, D, N)}}}
        \State Calculate softmax probability $p_i$. 
    \EndFor

    \vspace{0.2em}
    {\footnotesize\textcolor{DarkBlue}{\texttt{// Step2: Fuse selected candidates using dynamic weighted fusion.}}}
    \vspace{0.2em}
    
    \State Obtain $\mathcal{X}_{\text{selected}}$ using Eq.~\eqref{eq:threshold_selection}.\quad{\footnotesize\textcolor{teal}{\texttt{\# Selecting based on adaptive threshold \( \tau \)}}}    
    \State Compute $P_f$ using Eq.~\eqref{eq:combined_assign_fuse}.
           {\footnotesize\textcolor{teal}{\texttt{\# shape: (L,D,K)}}}

    \vspace{0.2em}
    {\footnotesize\textcolor{DarkBlue}{\texttt{// Step3: Training schedule.}}}
    \vspace{0.2em}

    \State Calculate feedback loss \( \mathcal{L}_{\text{feed}} \) using Eq.~\eqref{eq:square_loss}.
    \State Compute final loss \( \mathcal{L} \) using q.~\ref{eq:full_training_loss}\quad{\footnotesize\textcolor{teal}{\texttt{\# Combination of $\mathcal{L}_{\text{lm}}$, $\mathcal{L}_{\text{fuse}}$, and $\mathcal{L}_{\text{feed}}$}}}
    \State Update model parameters based on it.
\EndFor
\State \Return Trained $\mathcal{T}$.
\end{algorithmic}
\end{algorithm}